\documentclass{article}
\usepackage{spconf,amsmath,graphicx}
\usepackage{hyperref}
\usepackage{multirow}
\usepackage[usenames,table]{xcolor}
\usepackage{fancyhdr}
\title{Diagnostic Classification of Lung Nodules Using 3D Neural Networks}
 \name{Raunak Dey$^{\star}$ \qquad Zhongjie Lu$^{\dagger}$ \qquad Yi Hong$^{\star}$}

 \address{$^{\star}$ Department of Computer Science, University of Georgia, Athens, GA, USA \\
     $^{\dagger}$ First Affiliated Hospital, School of Medicine, Zhejiang University, Hangzhou, Zhejiang, China} 

\begin{document}
	%
	
	\maketitle
	
	\lhead{Accepted for publication in IEEE International Symposium on Biomedical Imaging (ISBI) 2018}  
	\rhead{Copyright $\copyright$ 2018 IEEE}

	\begin{abstract}
		
		Lung cancer is the leading cause of cancer-related death worldwide. Early diagnosis of pulmonary nodules in Computed Tomography (CT) chest scans provides an opportunity for designing effective treatment and making financial and care plans. In this paper, we consider the problem of diagnostic classification between benign and malignant lung nodules in CT images, which aims to learn a direct mapping from 3D images to class labels. To achieve this goal, four two-pathway Convolutional Neural Networks (CNN) are proposed, including a basic 3D CNN, a novel multi-output network, a 3D DenseNet, and an augmented 3D DenseNet with multi-outputs. These four networks are evaluated on the public LIDC-IDRI dataset and outperform most existing methods. In particular, the 3D multi-output DenseNet (MoDenseNet) achieves the state-of-the-art classification accuracy on the task of end-to-end lung nodule diagnosis. In addition, the networks pretrained on the LIDC-IDRI dataset can be further extended to handle smaller datasets using transfer learning. This is demonstrated on our dataset with encouraging prediction accuracy in lung nodule classification.
	\end{abstract}

	\section{Introduction}
	Lung cancer, the most common cause of cancer deaths, accounted for 1.69 million deaths worldwide in 2015. In 2018 there will be 234,030 estimated new cases diagnosed as lung and bronchus cancer and 154,050 estimated deaths in the United States~\cite{siegel2018cancer}. Such situation can be improved by using screening with low-dose computed tomography (CT), which has been shown to reduce lung cancer mortality~\cite{national2011reduced}.  
	However, due to the subtle differences between benign and malignant pulmonary nodules, lung cancer diagnosis is a difficult task even for human experts. In addition, the accuracy of radiologic diagnosis varies and greatly depends on the clinician's subjective experience. Computer-aided diagnosis (CAD) provides a non-invasive solution and an objective prediction for assisting radiologists in the lung nodule diagnosis. 
	
	Existing CAD methods fall into two categories: classification models based on hand-crafted features~\cite{krishnamurthy2016three,liu2016radiological,shewaye2016benign} and deep neural networks with automatic feature extraction~\cite{shen2015multi,nibali2017pulmonary,liu2017multiview,hussein2017risk,zhu2017deeplung}. Approaches in the first category typically measure radiological traits, e.g., nodule size, location, shape, texture, and adopt a classifier to determinate malignancy status. However, collecting and selecting a useful subset of features for lung nodule diagnosis is non-trivial and runs the risk of introducing measurement errors. In the second category, models based on deep neural networks can automatically learn features for diagnosis from lung CT images. They have shown promising prediction accuracy for risk stratification of lung nodules. However, most of existing deep learning models handle the pulmonary nodule classification problem by utilizing 2D convolutional neural networks (CNN)~\cite{shen2015multi} or multiview 2D CNNs to mimic 3D image volumes~\cite{nibali2017pulmonary,liu2017multiview}. They discard partial but important information of the original image data for classification. The model proposed in~\cite{hussein2017risk} uses 3D CNNs to extract features for six attributes of lung nodules and fuses those features to regress the malignancy score. In practice, providing those attributes of a nodule requires additional work from radiologists. Another recent model classifies lung nodules using gradient boosting machine with constructed features from 3D networks, nodule size, and raw nodule pixels~\cite{zhu2017deeplung}. 
	
	In this paper we propose four 3D neural networks for end-to-end lung nodule diagnosis, i.e., classifying a nodule in a 3D CT scan as benign or malignant. The basic network is a simple 3D CNN with two pathways accepting 3D image patches at different scales as input. The basic 3D CNN can be improved by providing intermediate outputs and/or adding connections between layers, which shorten the distance from input to output and could achieve better optimization results. Therefore, based on the basic 3D CNN we propose another three 3D networks, one with multiple intermediate outputs, one with dense blocks (a series of convolutional layers all connected to every other layer~\cite{huang2016densely}), and their combination including both intermediate outputs and dense connections.   
	
	\begin{figure*}[t]
		\centering
		\begin{tabular}{cc}
		\includegraphics[width=1.0\columnwidth]{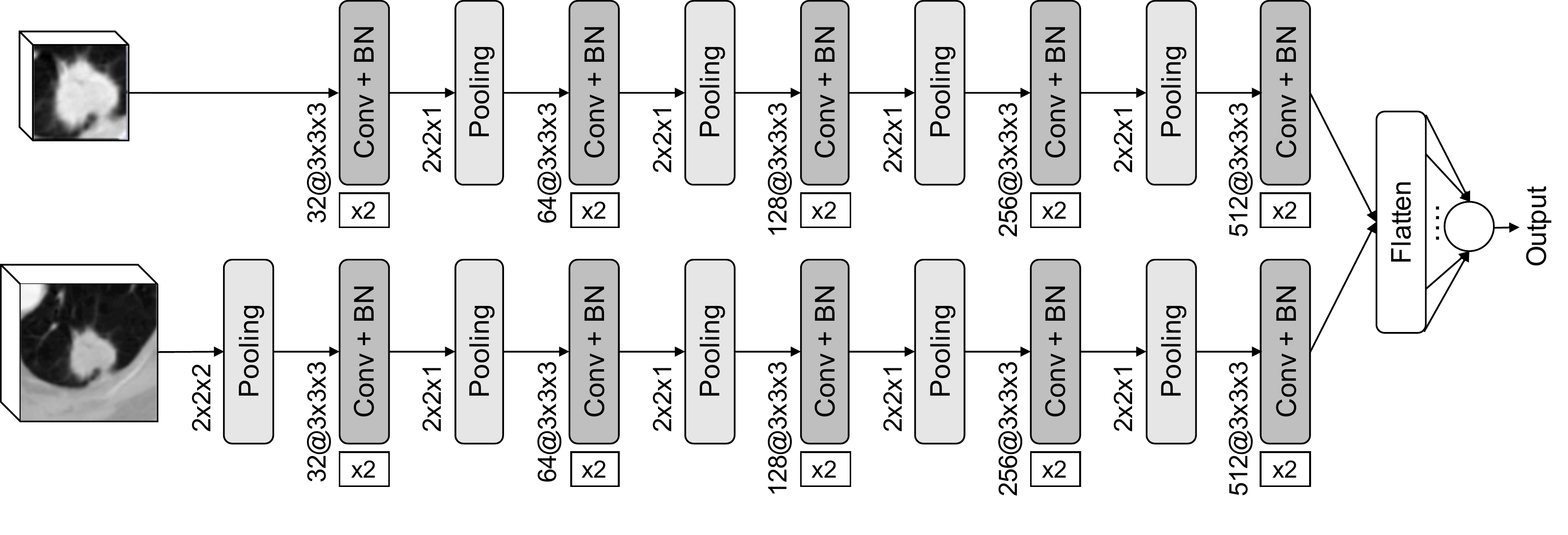} &
		\includegraphics[width=1.0\columnwidth]{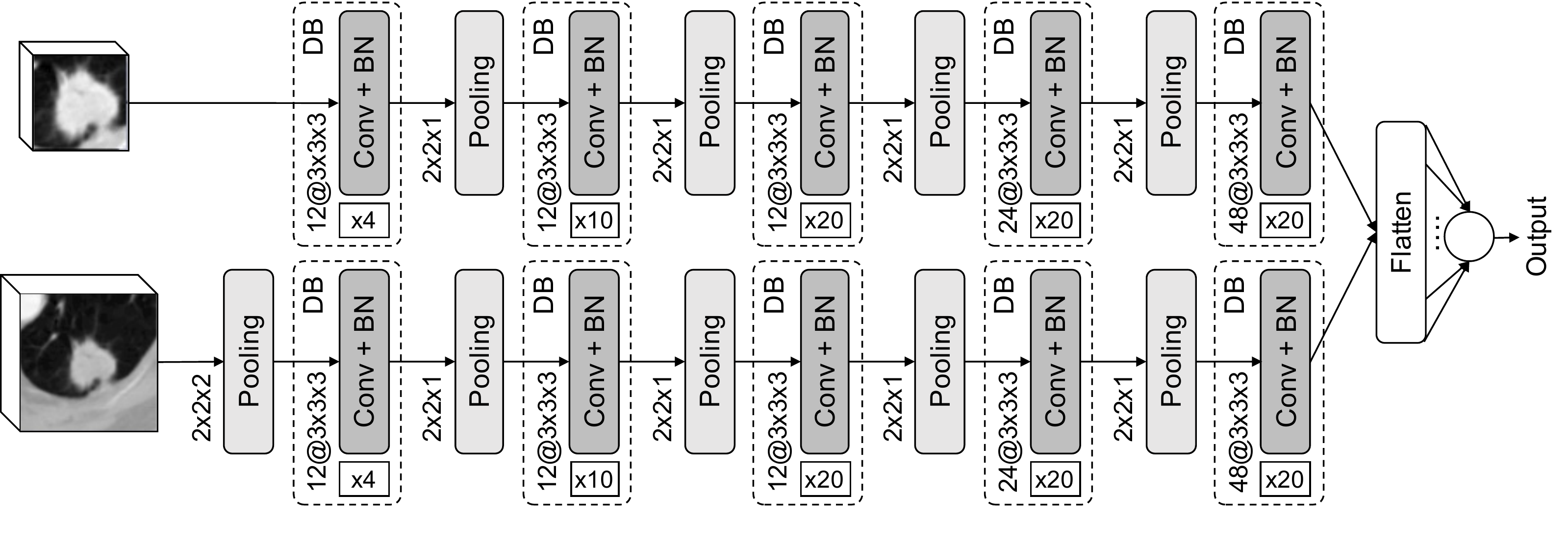}\\
		\small{(a) Basic 3D CNN} & \small{(c) 3D DenseNet} \\
		\quad \\
		\includegraphics[width=1.0\columnwidth]{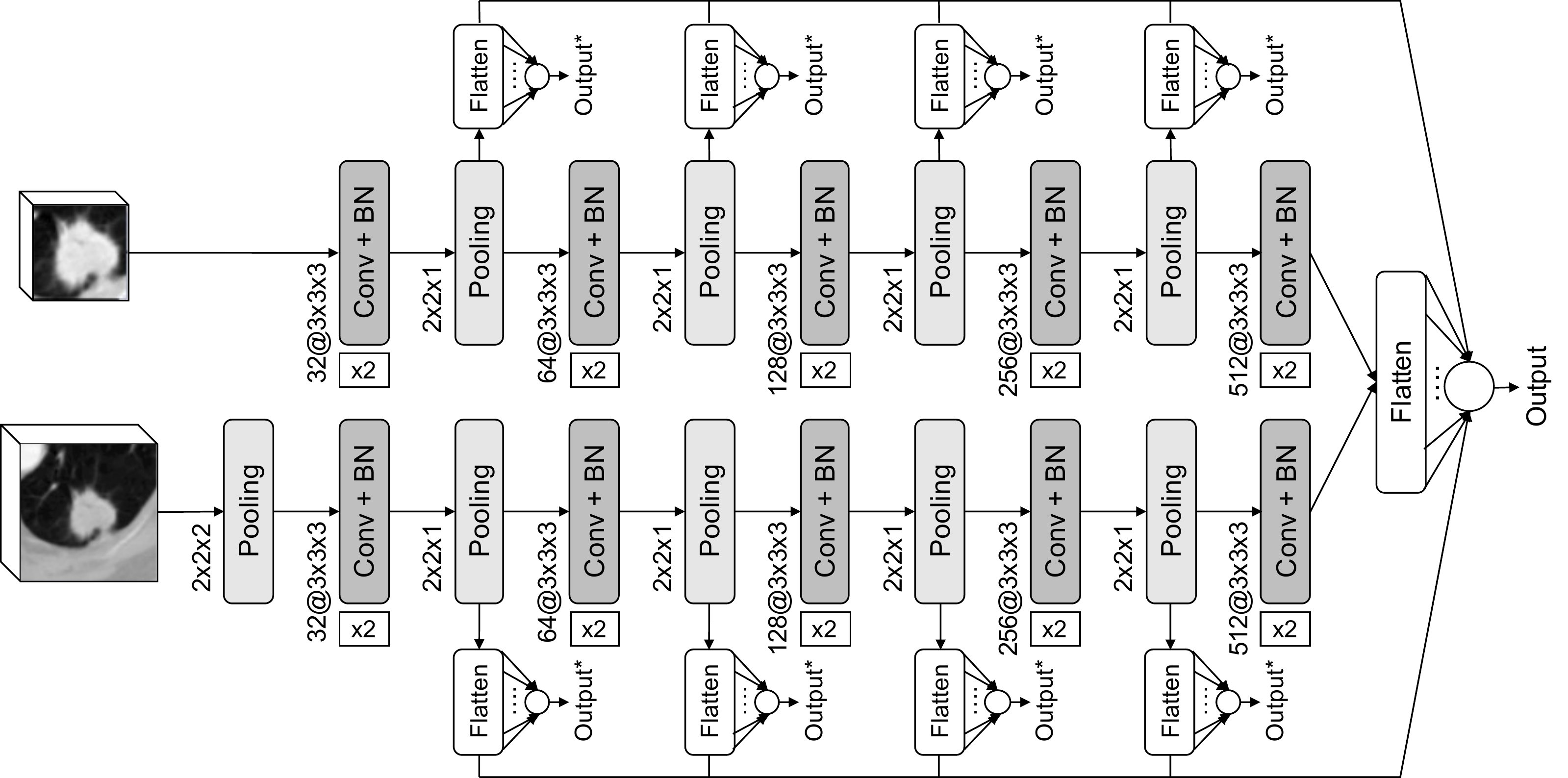}&
		\includegraphics[width=1.0\columnwidth]{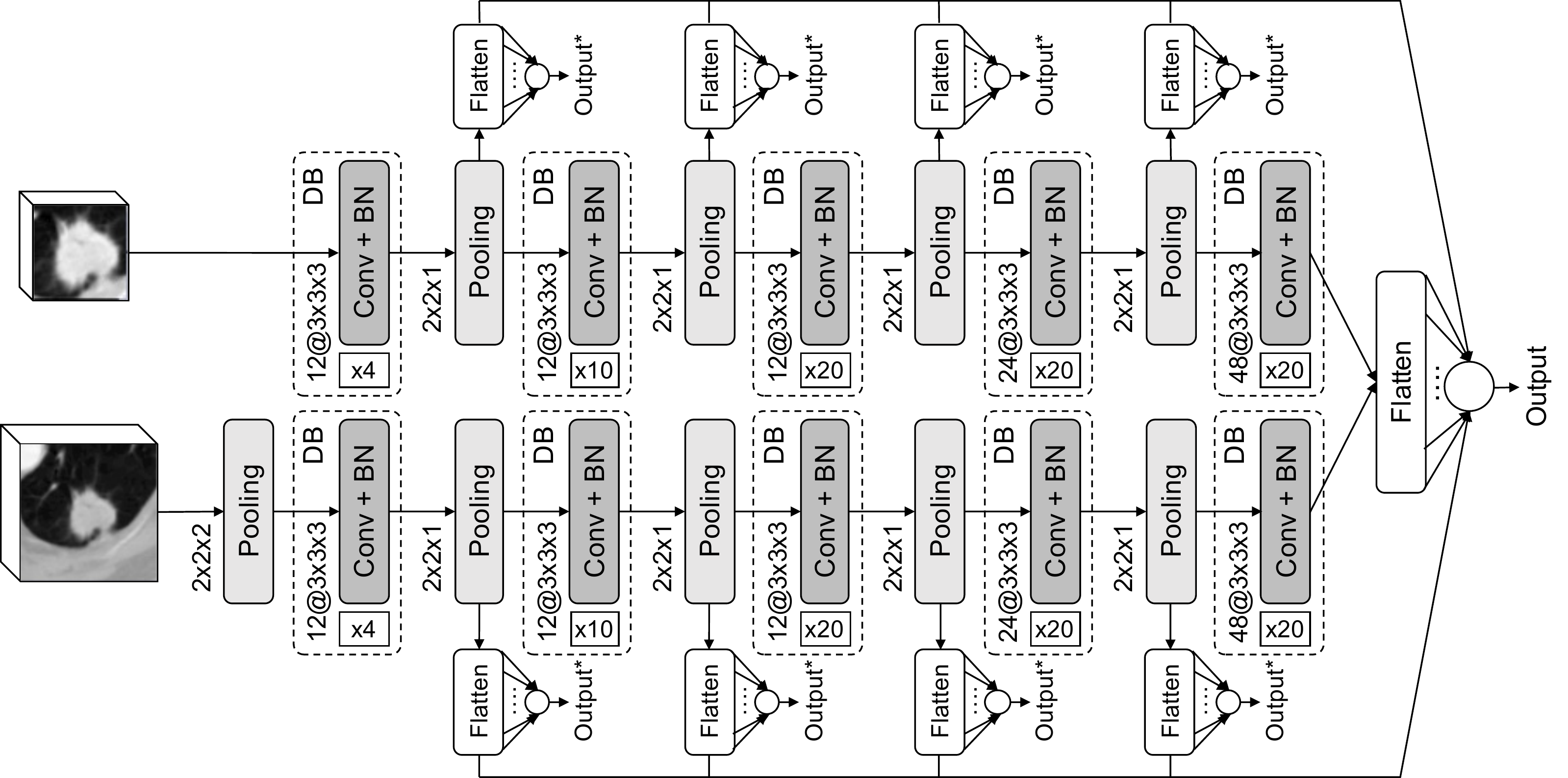}\\
		\small{(b) 3D Multi-Output CNN ($^\star$intermediate output)} & \small{(d) 3D Multi-Output DenseNet ($^\star$intermediate output)}
		\end{tabular}
		\caption{Architectures of our proposed networks. Conv: convolutional layer; BN: batch normalization; DB: dense block.}
		\label{fig:networks}
	\end{figure*}     
	
	The proposed networks are evaluated on the public LIDC-IDRI dataset~\cite{armato2011lung} and our private pulmonary nodule dataset. For the LIDC-IDRI dataset, the best performance is achieved by the 3D multi-output DenseNet (MoDenseNet), resulting in $90.40\%$ accuracy in a cross-validation setting. This outperforms a recent model for pulmonary nodule classification~\cite{nibali2017pulmonary}, by providing both higher classification accuracy and larger area under the receiver operating characteristic (ROC) curve. To the best of our knowledge, our proposed MoDenseNet achieves the state-of-the-art lung nodule classification result on the LIDC-IDRI dataset using an end-to-end model and 3D CT images with uniform slice thickness. In addition, for our private dataset the MoDenseNet also demonstrates the best performance among the four proposed 3D networks.

	\section{Methods}
	\label{sec:method}

	In this section we describe our classification models for lung nodules in CT scans using deep neural networks. In practice, radiologists make a diagnosis by checking multiple slices of a pulmonary nodule and considering 3D information of the nodule. Single- or multi-view 2D images as used in most of previous methods do not contain a complete 3D information for a lung nodule. Therefore, our proposed networks take as input a 3D CT chest scan with the location of a nodule and discriminate a malignant pulmonary nodule from benign ones. 
	
A typical CT scan consists of hundreds of 2D slices, e.g., a sequence of gray images with a dimension of $512 \times 512$. Directly dealing with the whole 3D image scan is not practical for a deep network due to limited memory resources. In addition, most lung nodules in our image data occupy less than 10 slices and some nodules are too small to be distinguished from surrounding normal tissues if viewing the full 3D image. Therefore, our networks focus on small regions centered at the annotated locations of lung nodules. The networks have two pathways that input 3D image patches at different scales to cover both local and global image context of a pulmonary nodule. With the aforementioned input and output, we next describe how to design the networks.

	\subsection{Network Architectures}
	\label{sec:networks}
	
	\noindent
	\textbf{Basic 3D CNN} \quad
	We first apply the basic principles of deep neural networks with 3D convolutional filters to the lung nodule classification problem. An accurate diagnosis requires both local detailed information of a lung nodule and global surrounding tissues for comparison. Therefore, our basic network has two pathways with 3D image inputs at two different scales, as shown in Fig.~\ref{fig:networks}(a). In particular, one pathway accepts a 3D image volume of dimension 50 pixel $\times$ 50 pixel $\times$ 5 slice, where a pulmonary nodule dominates. The other pathway accepts an image volume of dimension 100 $\times$ 100 $\times$ 10, which covers both a nodule and its surrounding tissues. The 3D patch size is selected according to our experimental datasets and might differ for a new dataset. 
	
	In the basic 3D network, the two pathways share the same structure, except that the one accepting a larger input has a max pooling layer with a filter of size $2 \times 2 \times 2$ before the first convolutional layer. Each pathway has ten convolutional layers, including a sequence of 32, 32, 64, 64, 128, 128, 256, 256, 512, and 512 feature maps generated by 3D convolutional filters of size $3 \times 3 \times 3$, and four max pooling layers with filters of size $2 \times 2 \times 1$, each after every two convolutional layers. Padding is used to maintain the size of feature maps after convolution operators and we apply batch normalization~\cite{ioffe2015batch} after every convolution layer. The feature maps from two pathways are concatenated and connected to the classification output layer. The depth of this network is adapted to the datasets used in the experiments and a deeper network may suffer from over fitting because of limited datasets.
		
\noindent
\textbf{3D Multi-Ouput CNN} \quad
	The basic 3D CNN would get stuck in a local optimum. To find a better optimum, we modify the basic 3D CNN by introducing early outputs, which provides immediate feedback from an early evaluation of error functions. Fig.~\ref{fig:networks}(b) demonstrates the network architecture modified from the basic 3D CNN by adding multiple outputs. Specifically, the network has intermediate outputs after every pooling layer that follows a convolutional layer. By collecting all feature maps before those intermediate outputs and those from the last convolutional layers in two pathways, we connect them to the classifier for the final output.  
	
	\noindent
	\textbf{3D DenseNet} \quad
	Similar to multi-output networks, DenseNet proposed in~\cite{huang2016densely} provides another way to shorten the distance from input to output. They can achieve a better optimization result, because the vanishing gradient is strengthened by the connections to layers closer to output. Therefore, our third network is adapted from DenseNet, which advances standard 2D neural networks by introducing dense blocks with direct connections among all the layers in a block. Hence, we modify the basic 3D CNN by introducing 3D dense blocks. The overall architecture shown in Fig.~\ref{fig:networks}(c) consists of five dense blocks for each pathway. The first three dense blocks consists of 4, 10, 20 convolutional layers, respectively. Each convolutional layer has 12 feature maps with 3D filters of size $3 \times 3 \times 3$. The last two dense blocks consist of 20 layers each and have 24 and 48 feature maps, respectively, which are also generated by 3D convolutional filters of size $3 \times 3 \times 3$.
	
	\noindent
	\textbf{3D Multi-Output DenseNet (MoDenseNet)} \quad
	Our final network adopts the design of the multi-output network to augment the 3D DenseNet. As shown in Fig.~\ref{fig:networks}(d), we follow the same strategy as used in the 3D multi-output CNN and provide early outputs after every pooling layer that follows the dense blocks in both pathways. The feature maps before each intermediate outputs are merged along with the features from the last convolutional layers of the two pathways and then sent to the classifier for the final output.
	
	\begin{figure}[t]
		\centering
		\begin{tabular}{cccc}
			\small{benign} & \small{malignant}  & \small{benign} & \small{malignant}\\
			\includegraphics[width=0.2\columnwidth]{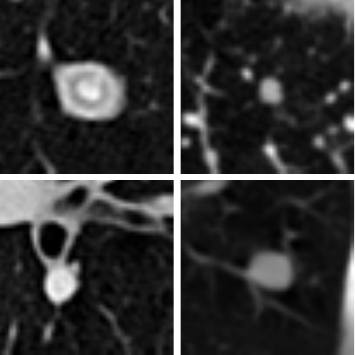} &
			\includegraphics[width=0.2\columnwidth]{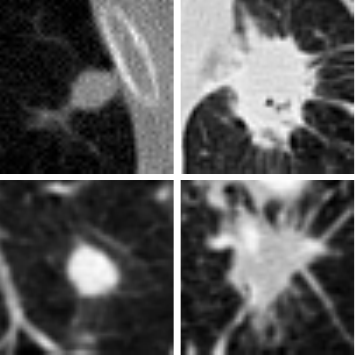} & 
			\includegraphics[width=0.2\columnwidth]{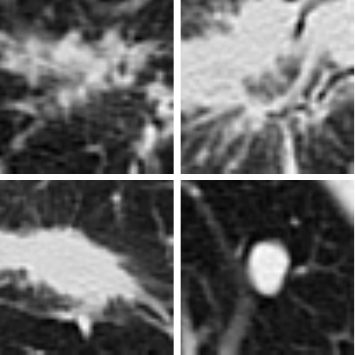} &
			\includegraphics[width=0.2\columnwidth]{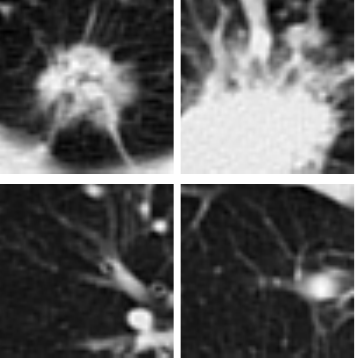}\\
			\multicolumn{2}{c}{\small{(a) LIDC-IDRI}} & \multicolumn{2}{c}{\small{(b) Our dataset}}
		\end{tabular}
		\caption{Pulmonary nodule samples from two datasets.}
		\label{fig:data_samples}
	\end{figure}
	
	\begin{table}[t]
		\centering 
		\small
		\begin{tabular}{c|ccccc} 
			\hline 
			3D Network & TPR\%  & TNR\% & PPV\% & AUC & ACC\% \\  
			\hline
			Basic CNN &  87.50  & 81.20  & 83.86 & 0.9064 &  84.35\\
			Multi-Output &  88.99  & 82.69  & 86.49 & 0.9205 & 85.84  \\
			DenseNet &  88.50  & 88.33  & 89.60 & 0.9451 &  88.42 \\	
			MoDenseNet & \cellcolor{green!10}{\bf 90.47} & \cellcolor{green!10}{\bf 90.33}  & \cellcolor{green!10}{\bf 90.55} & \cellcolor{green!10}{\bf 0.9548} & \cellcolor{green!10}{\bf 90.40}\\ 
			\hline
		\end{tabular}
		\caption{Results of the proposed networks on the LIDC-IDRI dataset. TPR: true positive rate or sensitivity; TNR: true negative rate or specificity; PPV: positive predictive value or precision; AUC: area under the ROC curve; ACC: accuracy, the percentage of correct predictions.} \label{table:lidc_results} 
	\end{table} 
	
	\subsection{Transfer Learning}
	\label{sec:transfer_leanring}
	
	To train the proposed networks, typically we need a large dataset. The LIDC-IDRI dataset has $686$ lung nodule samples available for experiments (see Sec.~\ref{sec:dataset} for more details). However, our dataset is relatively small and consists of only $147$ nodule samples. Therefore, for our dataset we augment our networks using transfer learning~\cite{bengio2012deep}. This is a common way to improve the performance of network on a smaller dataset by using a pretrained network on a similar but often larger dataset. In particular, we freeze all the  convolutional layers of the proposed networks that are pretrained on the LIDC-IDRI dataset and then retrain the last layer of each network on our private dataset.

	\section{Experiments}
	\label{sec:experiments}
	
	
	\subsection{Dataset and Preprocessing}
	\label{sec:dataset}
	
	\noindent
	\textbf{LIDC-IDRI dataset} \quad
	This public dataset consists of 1010 CT scans with annotations provided by four radiologists. Each image includes one or more lung nodules. We use the nodule list reported in~\cite{Reeves2011lidc} to obtain the x, y coordinate and slice number of a nodule's location, which is used to generate 3D image patches as input for the networks. To collect the diagnosis information of nodules, we follow the preprocessing steps in~\cite{nibali2017pulmonary}. Each nodule is annotated by one to four radiologists and given a grade from 1 to 5, with 1 and 5 being the extremes of benign and malignancy, respectively. A zero grade indicates unavailable diagnosis~\cite{zhu2017deeplung}, which will be ignored. We consider lung nodules diagnosed by at least 3 radiologists, compute the median value of annotated grades for a nodule, and take a median value of less than 3 as benign and greater than 3 as malignant. We exclude all nodules with a median value equal to 3. This results in a collection of $808$ lung nodules with unique annotations. Furthermore, some lung nodules are discarded because of missing slices and/or non-uniform slice thickness. At the end, we have $686$ lung nodules ({46\% benign and 54\% malignant}) for experiments.

	\begin{table}[t]
		\centering 
		\small
		\begin{tabular}{c|cccc} 
			\hline 
			\multirow{2}{*}{3D Network} &\multicolumn{2}{c}{W/O Transfer Learning}&\multicolumn{2}{c}{W/ Transfer Learning} \\  
			& AUC & ACC\% & AUC & ACC\%\\
			\hline
			Basic CNN & 0.6547  &  60.21 & 0.7087 & 64.54 \\
			Multi-Output & 0.7537  &  70.90 & 0.7702 & 74.54 \\
			DenseNet &  0.8018 & 81.21  & 0.8468 & 83.63\\ 
			MoDenseNet & 0.8634 & 85.45 & \cellcolor{green!10}{\bf 0.9010} & \cellcolor{green!10}{\bf 86.84}\\ 
			\hline 
		\end{tabular}
		\caption{Results of the proposed networks on our dataset with and without transfer learning. AUC: area under the ROC curve; ACC: accuracy, the percentage of correct predictions.} \label{table:our_results} 
	\end{table} 
	
	\noindent
	\textbf{Our dataset} \quad
	We have collected $147$ CT scans ({37\% benign and 63\% malignant}) for this lung cancer diagnosis problem. The location for each pulmonary nodule is specified by the radiologist and each nodule was determined whether it is benign or malignant by taking a biopsy of the nodule. Compared to the LIDC-IDRI dataset, our dataset is even more challenging with more subtle differences between benign and malignant lung nodules as shown in Fig.~\ref{fig:data_samples}. 
	
	\subsection{Experiment Settings}
	\label{sec:settings}	
	The implementation of our networks is based on Keras with tensorflow as backend~\cite{chollet2015keras}. Since our classification models are of a binary nature and deal with unbalanced datasets, we use the weighted binary cross entropy as the loss function. The LIDC-IDRI dataset is evaluated using 5-fold cross-validation, four subsets for training/validation and one for testing. Since we have limited data samples for training, we only use $2.5\%$ of the four subsets for validation. All proposed networks are trained using Adadelta optimizer and L2 regularizer. For the LIDC-IDRI dataset, networks are trained from scratch; while for our dataset, the networks are pretrained on the whole available LIDC-IDRI dataset, with $10\%$ random samples used in the validation set. Because the sample size of our dataset is smaller than that of the LIDC-IDRI dataset, we use 3-fold cross-validation in the experiments, one subset for testing, $5\%$ of the two subsets for validation, and the remaining data samples for training. The maximum iteration for training from scratch is set to 150 epochs, while in the transfer learning phase, we retrain the last fully-connected layer of the networks for 20 more epochs on our private training dataset.

	\subsection{Experimental Results}
	\label{sec:results}
	
\begin{figure}[t]
		\centering
		\includegraphics[height=0.45\columnwidth]{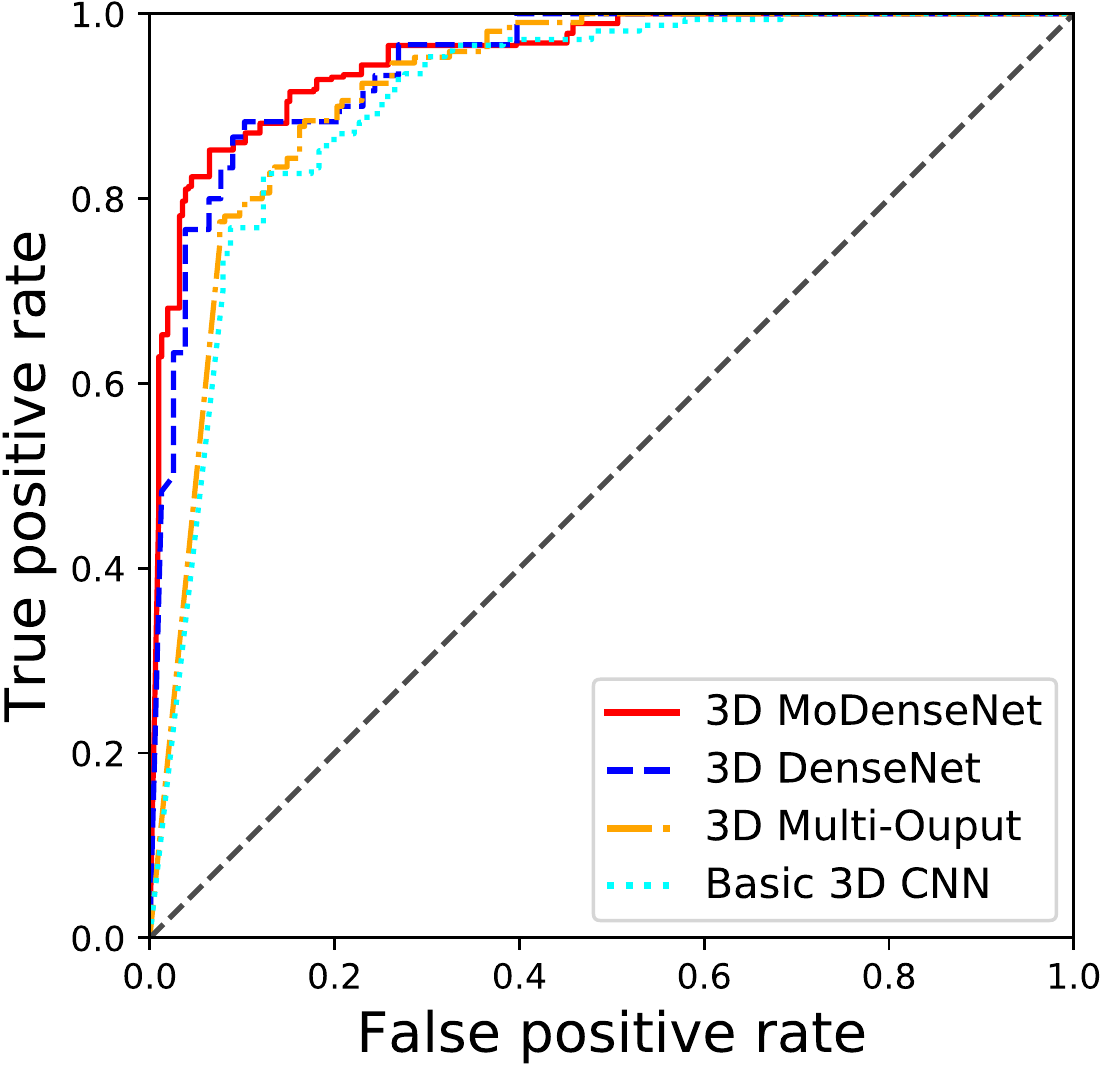} \quad
		\includegraphics[height=0.45\columnwidth]{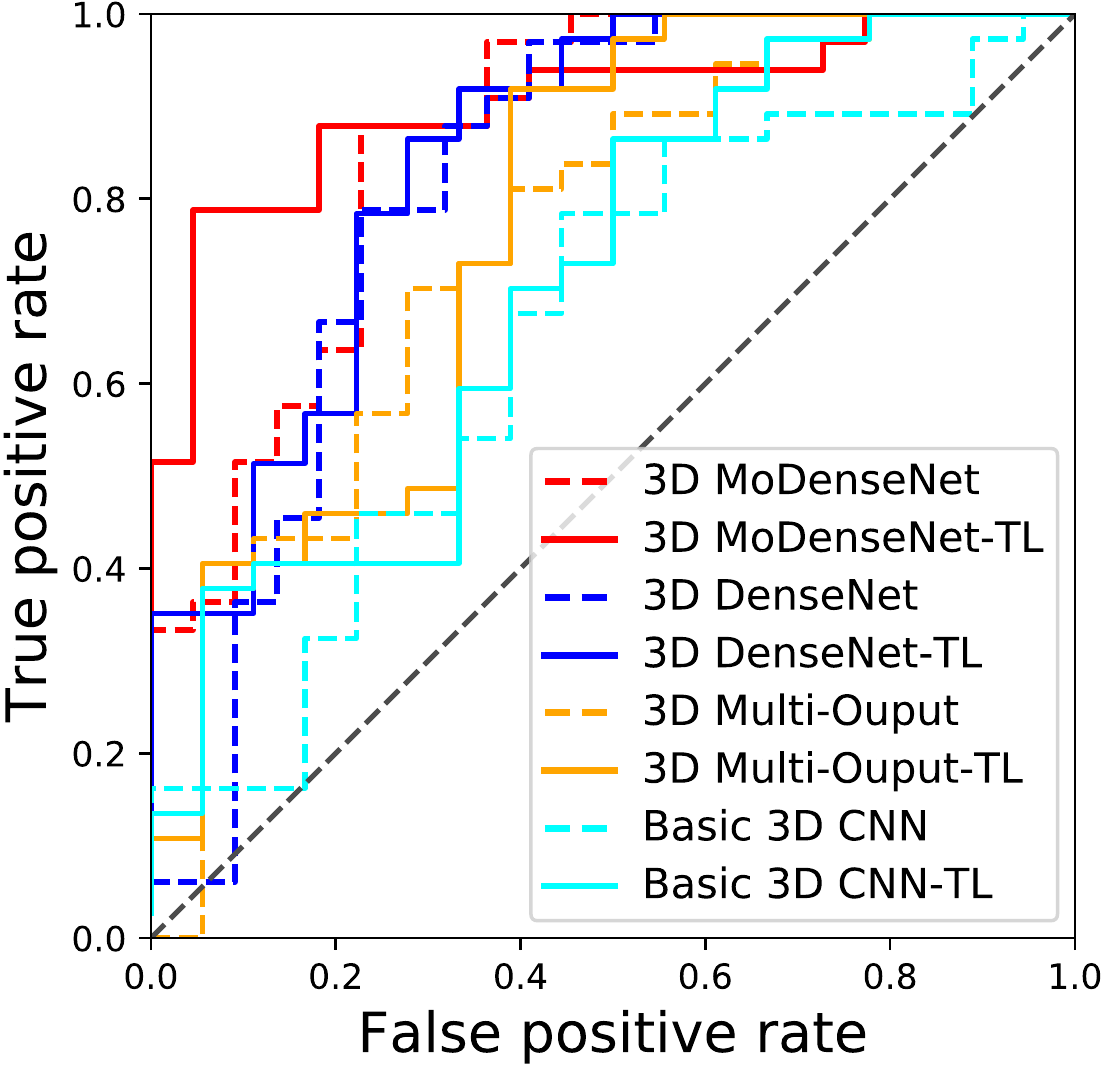}
		\caption{ROC curves for our proposed networks tested on the LIDC-IDRI dataset (left) and our dataset (right) with and without transfer learning (TL).}
		\label{fig:roc}
	\end{figure}

	Table~\ref{table:lidc_results} shows the experimental results of our proposed networks on the LIDC-IDRI dataset and their ROC curves are shown in Fig.~\ref{fig:roc} (left). The multi-output DenseNet obtains the highest accuracy of $90.40\%$ and area under the curves (AUC) of $0.9548$. In addition, we estimate the model size and the number of parameters for these four networks (basic 3D CNN, 3D multi-output net, 3D DenseNet, and 3D MoDenseNet) are $28$, $29$, $34.6$, and $34.8$ million, respectively. The results indicate that although DenseNet and multi-output networks have complicated architectures with more parameters compared to the basic 3D CNN, the approach of shortening distance between input and output, with early outputs and/or dense connections among convolutional layers, does benefit the optimization and helps achieve better optimal solutions.
	
	For comparison, the most related work~\cite{nibali2017pulmonary} handled a similar subset of the LIDC-IDRI with 831 lung nodules and reported $0.9459$ AUC and $89.90\%$ accuracy. Their result was computed based on $20\%$ selected nodule data for testing and ours are evaluated in a cross-validation setting which is more rigorous. Under this condition, our multi-output DenseNet still outperforms their method in terms of prediction accuracy, AUC, and all other metrics used in Table~\ref{table:lidc_results}, except for the sensitivity (ours: $90.47\%$ and theirs: $91.07\%$). 
	
	On our dataset we apply the four proposed networks and they all demonstrate improved area under the ROC curves (Fig.~\ref{fig:roc}, right) and classification accuracy by using transfer learning (see Table~\ref{table:our_results}). The performance ranking of the four networks on our dataset keeps the same with that on the LIDC-IDRI dataset, i.e., the 3D MoDenseNet performs best, followed by 3D DenseNet, 3D multi-output network, and then the basic 3D CNN, either with or without transfer learning. Compared to the LIDC-IDRI dataset, this dataset is more challenging and the radiologist claims around $70\%$ accuracy of diagnosis by simply checking CT scans. Our results are encouraging and the classification accuracy reaches to $86.84\%$. Overall, the 3D multi-output DenseNet demonstrates consistent promising performance on both datasets.
	
%
	
	\section{Discussion}
	\label{sec:discussion}
	In this paper, we investigated 3D networks to classify pulmonary nodules in a CT image into benign or malignant categories. Compared to using 2D slices or approximating 3D image with multi-views, directly working on 3D volumes yields better results for the lung nodule classification problem when the slice thickness is consistent. In addition, the optimization of a network is more efficient by having a prompt error back-propagation. This can be achieved by adding connections between layers or by introducing early outputs. 
	
	Our experimental results demonstrate decent classification performance in the lung cancer diagnosis. Next, we aim at understanding the features extracted by the networks for classification using various visualization techniques and at determining whether those features are consistent with features used by radiologists for diagnosis. Another future work is automatic pulmonary nodule detection, which will relax the requirement of manual annotations for nodule locations.

	\bibliographystyle{IEEEbib}
	\bibliography{lung_nodule_classification}

\end{document}